\documentclass{article}
\usepackage[preprint]{colm2026_conference}

\usepackage{microtype}
\usepackage{hyperref}
\usepackage{url}
\usepackage{booktabs}
\usepackage{xcolor}
\usepackage{graphicx}
\usepackage{tikz}
\usepackage{amsmath}
\usepackage{lineno}
\usetikzlibrary{positioning}

\usepackage{amsmath,amsfonts,bm}

\def\eqref#1{equation~\ref{#1}}

\def\1{\bm{1}}

\DeclareMathAlphabet{\mathsfit}{\encodingdefault}{\sfdefault}{m}{sl}
\SetMathAlphabet{\mathsfit}{bold}{\encodingdefault}{\sfdefault}{bx}{n}

\definecolor{darkblue}{rgb}{0, 0, 0.5}
\hypersetup{colorlinks=true, citecolor=darkblue, linkcolor=darkblue, urlcolor=darkblue}

\title{When Should Forecasting Agents Reason?\\Behavioral Stress Tests for Reliability Routing}

\author{Louis Wang}

\newcommand{\tax}{\textsc{TaxonomyRoute}}
\newcommand{\rel}{\textsc{ReliabilityRoute}}
\newcommand{\fitroute}{\textsc{FittedRoute}}
\newcommand{\hist}{\textsc{HistoricalAnalog}}
\newcommand{\search}{\textsc{SearchLoop}}
\newcommand{\aialocal}{\textsc{AIA-Repro}}

\begin{document}

\ifcolmsubmission
\linenumbers
\fi

\maketitle

\begin{abstract}
Forecasting agents increasingly combine language-model reasoning, retrieval, ensembling, and calibration, but it remains unclear when each behavior should be trusted. We study this question on ForecastBench-style binary forecasting tasks, treating the choice to retrieve, reason, defer to a market prior, or use a historical analog as an observable agent behavior rather than a hidden implementation detail. Our central finding is that mechanism choice is source-dependent: structured analogs dominate for some data-generating processes, while market/crowd-style and conservative baselines are better for others. We introduce \rel{}, a structural intervention that steers forecasting-agent behavior using reliability features such as historical coverage, market-prior availability, source-prior sharpness, evidence strength, evidence disagreement, and horizon. A fixed 2024-fitted rule closely matches a hand taxonomy without hard-coded source-name decisions, while a walk-forward self-adjusting rule refits thresholds from previously resolved vintages and obtains the best mean Brier score among our deterministic systems across 16 later LLM vintages. The gain is modest and historical/search baselines remain highly competitive. The main contribution is therefore a behavioral stress test showing that more reasoning is not always better; forecasting agents should first estimate which evidence source deserves control, routing policies should themselves adapt under auditable constraints, and reproducibility artifacts are available at \url{https://github.com/louiswang524/forcastagent}.
\end{abstract}

\section{Introduction}

Forecasting agents estimate probabilities for future events from heterogeneous evidence such as time-series histories, public forecasts, market prices, and textual context. This setting is important because forecasts guide decisions under uncertainty, and benchmarks such as ForecastBench provide a dynamic way to evaluate probabilistic forecasting systems without relying on already-resolved static datasets \citep{karger2024forecastbench}. Many recent systems combine retrieval, multiple reasoning roles, supervisor aggregation, and calibration. This leaves a behavioral design question open: should every forecasting question trigger the same agentic reasoning behavior, or should the system first decide whether to reason, retrieve, defer to a prior, or use a structured analog?

The core behavioral challenge is that forecasting evidence is not exchangeable across sources. A macroeconomic time-series question may be best answered by leakage-checked historical analogs, a prediction-market question may already contain a strong crowd prior, and an open-ended news question may require retrieval and textual reasoning. Treating these settings with one uniform ensemble creates an over-reasoning failure mode: the agent invokes expensive or brittle reasoning behavior even when a simpler evidence source is more reliable. Conversely, selecting the best component on a calibration set can overfit the mixture of sources and horizons in that vintage. The forecasting-agent design problem is therefore not only how to build better reasoners, but how to steer behavioral mechanism choice without leaking evaluation labels.

We propose to study forecasting agents as reliability-routed systems. Instead of presenting another monolithic agent, we build a reproducible harness that decomposes a forecasting system into four mechanisms: a reproduced multi-agent forecasting baseline, an evidence-graph search loop, a leakage-checked historical analog forecaster, and feature-based routing over these mechanisms. Our key method, \rel{}, fits a small threshold rule over non-label reliability features rather than choosing per-source winners. We evaluate both a frozen rule, which tests direct transfer from one calibration vintage, and a self-adjusting rule, which updates thresholds only from previously resolved vintages in a walk-forward protocol.

Our experiments produce three findings. First, fixed reliability routing can improve over the best single deterministic baseline on a later ForecastBench vintage, reducing Brier score from 0.1876 for \hist{} to 0.1846 for \rel{} on the 2025-10-26 evaluation set. Second, fixed thresholds do not uniformly transfer: across 16 LLM vintages, \hist{} and \search{} remain highly competitive. Third, walk-forward self-adjustment closes this gap, reducing mean Brier from 0.1867 for the frozen reliability router to 0.1839 and beating the frozen rule on 10 of 16 vintages. These results suggest that a strong forecasting agent should not simply reason more. It should decide which kind of evidence deserves control, while treating routing policies themselves as adaptive behaviors that must be stress-tested across vintages.

Our contributions are:
\begin{itemize}
    \item We introduce a reproducible behavioral stress-test harness for ForecastBench-style forecasting agents, treating mechanism choice as an observable behavior alongside probabilistic accuracy.
    \item We introduce \rel{}, a feature-based reliability router that replaces hard-coded source-name routing with auditable thresholds over historical coverage, market availability, prior sharpness, evidence strength, disagreement, and horizon.
    \item We show that fixed reliability routing can outperform the best single deterministic baseline on one later vintage, and that walk-forward self-adjustment improves average multi-vintage performance while preserving an inspectable rule family.
    \item We provide behavioral route diagnostics, source-slice diagnostics, threshold-stability analysis, multi-vintage stress tests, oracle-routing headroom, and calibration ablations showing when structured analogs, market/crowd priors, and agentic reasoning are useful.
\end{itemize}

\section{Method}
\label{sec:method}

\subsection{Problem Setting}

We consider binary forecasting questions. For each question $q$, a system outputs a probability $p \in [0,1]$ that the event resolves true. Once the event resolves with outcome $y \in \{0,1\}$, we evaluate forecasts with Brier score $(p-y)^2$ \citep{brier1950verification}, log score, and calibration error. Lower Brier and log score are better. Each ForecastBench target also has metadata such as source family, forecast due date, resolution date, and, when available, a frozen market or crowd probability.

\subsection{Behavioral Mechanisms}

Our harness separates forecasting into behavioral mechanisms rather than treating the agent as a single opaque policy. This decomposition is the main experimental object: it lets us ask not only whether an agent is accurate, but what kind of behavior produced the forecast.

\textbf{Historical analog forecaster.} For structured sources such as FRED, DBnomics, and YFinance, we build leakage-checked analog probabilities from observations dated no later than the forecast due date. For each target, the forecaster computes the resolution horizon in days, scans historical observations before the cutoff, and records whether the value increased over matching horizons. A series is used only when at least 20 comparable historical outcomes are available. The final probability is a clipped blend of the historical increase rate, a neutral prior, and a small mean-reversion adjustment based on the frozen value's historical percentile. If no historical analog is available, the forecaster falls back to the source prior embedded in the question metadata.

\textbf{Search and evidence-graph forecaster.} The search loop plans four specialist queries: base rate, recent evidence, resolution criteria, and counterevidence. The local search provider retrieves only frozen benchmark evidence and leakage-checked historical analog documents; it does not call live web search in the reported deterministic experiments. A deterministic reasoner then updates a logit prior from retrieved document weights, historical-analog probabilities, and market/crowd probabilities, and aggregates role forecasts by confidence-weighted averaging in logit space. This module approximates the structure of LLM search agents while keeping the ablation deterministic and inexpensive.

\textbf{Reproduced multi-agent baseline.} We implement an AIA-style reproduced baseline for comparison \citep{alur2025aia}: five focused agents forecast from different evidence groups, a supervisor aggregates their probabilities using a median, and a fixed extremization slope of 1.15 is applied in logit space. This baseline is not meant to be a claim about the original system's exact performance; it is a reproducible comparator that captures the broad motifs of role-specialized forecasting, supervisor aggregation, and extremization.

\subsection{Reliability Routing}

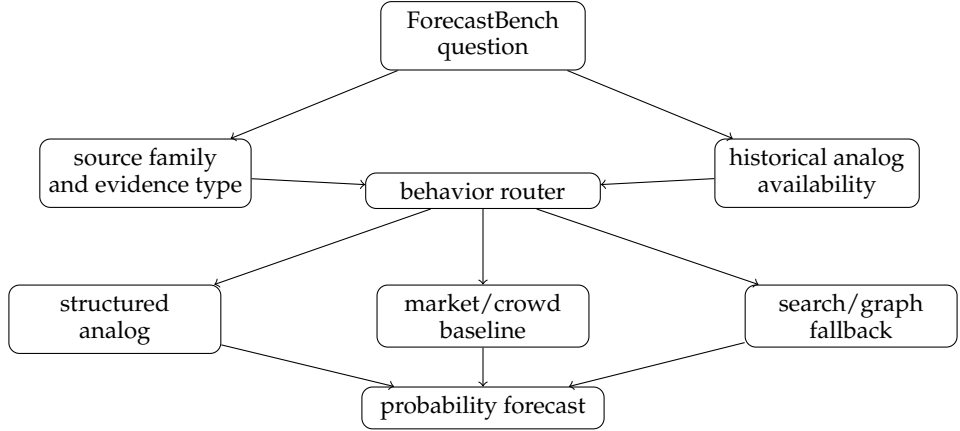
\begin{figure}[t]
\centering
\begin{tikzpicture}[node distance=1.05cm, every node/.style={font=\small}]
  \node[draw, rounded corners, align=center, minimum width=2.7cm, minimum height=0.75cm] (q) {ForecastBench\\question};
  \node[draw, rounded corners, align=center, below left=0.9cm and 1.7cm of q, minimum width=2.7cm] (meta) {source family\\and evidence type};
  \node[draw, rounded corners, align=center, below right=0.9cm and 1.7cm of q, minimum width=2.7cm] (hist) {historical analog\\availability};
  \node[draw, rounded corners, align=center, below=1.35cm of q, minimum width=3.1cm] (router) {behavior router};
  \node[draw, rounded corners, align=center, below left=1.0cm and 1.9cm of router, minimum width=2.8cm] (analog) {structured\\analog};
  \node[draw, rounded corners, align=center, below=1.0cm of router, minimum width=2.8cm] (aia) {market/crowd\\baseline};
  \node[draw, rounded corners, align=center, below right=1.0cm and 1.9cm of router, minimum width=2.8cm] (search) {search/graph\\fallback};
  \node[draw, rounded corners, align=center, below=2.35cm of router, minimum width=3.2cm] (p) {probability forecast};
  \draw[->] (q) -- (meta);
  \draw[->] (q) -- (hist);
  \draw[->] (meta) -- (router);
  \draw[->] (hist) -- (router);
  \draw[->] (router) -- (analog);
  \draw[->] (router) -- (aia);
  \draw[->] (router) -- (search);
  \draw[->] (analog) -- (p);
  \draw[->] (aia) -- (p);
  \draw[->] (search) -- (p);
\end{tikzpicture}
\caption{\rel{} routes each forecasting target to a mechanism based on reliability features rather than fitting per-source winners on the evaluation distribution.}
\label{fig:pipeline}
\end{figure}

We compare four behavioral routing policies. These are internal baselines and diagnostics, not public named systems. \textbf{Analog gating} uses a pure historical analog probability whenever the search provider retrieves one; otherwise, it uses search/graph fallback. \textbf{\fitroute{}} is a label-fit diagnostic: for each source family, it chooses the mechanism with the best Brier score on the 2024 calibration vintage, then freezes that source-to-mechanism mapping for evaluation. It is included to test whether direct empirical source fitting transfers, not as a deployable method. \textbf{\tax{}} is a hand-written diagnostic router that maps broad source families to mechanisms using source semantics: structured historical sources are routed to historical analogs, market/crowd sources to the reproduced multi-agent baseline, and unsupported cases to search/graph fallback. It provides an interpretable reference point for how much of routing can be explained by obvious source taxonomy. Our proposed method, \textbf{\rel{}}, removes source-name decisions and fits a small reliability rule over non-label features. Figure~\ref{fig:pipeline} summarizes this routing view.

\begin{table}[t]
\centering
\caption{Initial \rel{} feature rule fitted on the 2024 calibration vintage. In the walk-forward variant, the same threshold family is refit from previously resolved vintages before each later evaluation vintage.}
\label{tab:reliability}
\begin{tabular}{@{}p{0.27\linewidth}p{0.63\linewidth}@{}}
\toprule
Feature or threshold & Role in routing \\
\midrule
Historical coverage $\ge 0.5$ & Allows historical analog routing when the target has an analog and the source has broad analog coverage \\
Market sharpness $\ge 0.05$ & Routes to \aialocal{} when a non-neutral market/crowd prior is available \\
Source prior sharpness $\ge 0.18$ & Routes to \hist{} fallback when source priors are informative even without explicit analogs \\
Evidence strength $\ge 0.30$ for historical sources & Treats high-strength structured evidence as noisy and routes to conservative \aialocal{} \\
Horizon $\le 400$ days & Prevents long-horizon analogs from controlling when the historical comparison is weak \\
Search strength $\ge 0.50$ and disagreement $\le 0.35$ & Allows search fallback only when retrieved evidence is strong and not internally conflicted \\
\bottomrule
\end{tabular}
\end{table}

The distinction between \fitroute{} and \rel{} is deliberate. A fitted source router can exploit calibration labels but may learn source-vintage artifacts. A reliability router is less flexible: it can only use historical coverage, market/crowd prior availability, prior sharpness, evidence strength, evidence disagreement, and horizon. We instantiate this rule family in two ways. The fixed router selects thresholds on the 2024 calibration vintage and freezes them. The self-adjusting router augments the semantic threshold grid with quantiles from the available calibration feature distribution and, in the multi-vintage protocol, refits thresholds before each evaluation vintage using only the original calibration set and earlier resolved vintages. The feature set, rule order, candidate-threshold construction, and component forecasters are fixed before walk-forward scoring begins; resolved feedback can only choose threshold values within this pre-specified family. This constraint turns routing into a behavioral intervention: it changes when the agent reasons, retrieves, or defers, while keeping the intervention inspectable.

\section{Experimental Setup}
\label{sec:setup}

\paragraph{Benchmark.} We use public ForecastBench dataset releases \citep{karger2024forecastbench}. Our primary transfer protocol calibrates design choices on the 2024-07-21 human-written vintage and evaluates without retuning on the 2025-10-26 LLM-written vintage. We also run a multi-vintage stress test on 16 later LLM-written ForecastBench question sets from 2025-03-02 through 2025-12-07, excluding combination questions that do not match our binary single-question protocol. For the self-adjusting router, each vintage is scored in chronological order after refitting thresholds on only previously resolved vintages. This tests whether a design rule survives changes in question provenance, source mixture, and resolution progress without using current-vintage labels.

\paragraph{Baselines.} We compare \rel{} against \tax{}, \hist{}, historical-evidence blending, \search{}, \fitroute{}, \aialocal{}, an evidence-graph baseline, and a market-only baseline where applicable. \tax{} and \fitroute{} are diagnostic routers introduced in this paper: \tax{} uses human-written source-family rules, while \fitroute{} uses calibration-label performance to choose one mechanism per source. \aialocal{} is our deterministic reproduction of the main architectural pattern in AIA Forecaster \citep{alur2025aia}; because the original system's full prompts, model choices, search stack, and calibration details are not all available in our harness, we use it as a baseline for comparison rather than as a leaderboard claim about the external system. We also include two diagnostic upper bounds: an oracle source router that chooses the best mechanism per source using 2025 labels, and an oracle question router that chooses the best mechanism per question using 2025 labels. These oracle systems are not deployable; they measure remaining routing headroom.

\paragraph{Behavioral and outcome metrics.} We report Brier score, log score, calibration error, and routing behavior. Brier and log score are strictly proper scoring rules for probabilistic forecasts \citep{gneiting2007strictly}. Calibration error is computed by binning predictions and comparing average predicted probability to empirical frequency. Routing behavior is measured by the mechanism selected for each target: historical analog use, conservative market/crowd-style baseline use, search-loop use, or graph-reasoning use. For paired comparisons, we bootstrap question IDs with replacement using 2000 samples and report 95\% percentile intervals for routed-system differences. For behavioral robustness, we perturb each fitted reliability threshold and measure both route-change rate and change in Brier score; for self-adjustment, we additionally log the active rule before every vintage.

\paragraph{Calibration ablation.} To test whether the routed system merely needs post-hoc calibration, we fit Platt-style transformations on the 2024 calibration vintage. A global calibration fits one slope and intercept in logit space. A source-conditional calibration fits source-specific slopes and intercepts when at least 20 calibration targets are available and otherwise falls back to the global fit. Both transformations are frozen before scoring the 2025 evaluation set.

\section{Results}
\label{sec:results}

Tables~\ref{tab:main}--\ref{tab:behavior} and Table~\ref{tab:multivintage} answer different questions. The first two tables evaluate direct transfer on one diagnostic vintage, including the frozen rule fitted only on 2024 data; the multi-vintage table evaluates chronological adaptation across 16 later vintages, where the self-adjusting rule may use only feedback from already resolved earlier vintages.

\subsection{No-Retune Transfer}

\begin{table}[t]
\centering
\caption{Single-vintage transfer result on 2025-10-26. The fixed row uses only the 2024 calibration vintage; the self-adjusting row uses the walk-forward rule available by this vintage. Lower is better for all metrics.}
\label{tab:main}
\begin{tabular}{lccc}
\toprule
System & Brier $\downarrow$ & Log score $\downarrow$ & Cal. error $\downarrow$ \\
\midrule
\tax{} & \textbf{0.1846} & \textbf{0.5393} & 0.0650 \\
\rel{} fixed & \textbf{0.1846} & \textbf{0.5394} & 0.0649 \\
\hist{} & 0.1876 & 0.5460 & 0.0240 \\
\rel{} self-adjusting & 0.1876 & 0.5461 & 0.0232 \\
Historical-evidence blend & 0.1886 & 0.5505 & \textbf{0.0231} \\
\search{} & 0.1888 & 0.5524 & 0.0271 \\
Analog-gated search & 0.1888 & 0.5524 & 0.0271 \\
\fitroute{} & 0.1949 & 0.5651 & 0.0797 \\
\aialocal{} & 0.2009 & 0.5808 & 0.1200 \\
Evidence graph & 0.2048 & 0.5858 & 0.1134 \\
Market only & 0.2296 & 0.6406 & 0.1180 \\
\bottomrule
\end{tabular}
\end{table}

Table~\ref{tab:main} shows that the fixed \rel{} rule matches the best Brier score among our deterministic non-oracle systems while using feature thresholds rather than source-name routing. It improves Brier by 0.0030 over the strongest single baseline, \hist{}, and by 0.0042 over the search loop. The gain is modest in absolute terms; paired-bootstrap intervals are $[-0.0062, 0.0005]$ against \hist{} and $[-0.0078, -0.0006]$ against \search{}. The self-adjusting rule is not better on this individual vintage, but it is much better calibrated and becomes stronger in the 16-vintage walk-forward test below. This distinction is important: fixed transfer asks whether one vintage's thresholds generalize, while walk-forward adaptation asks whether the routing behavior can improve as resolved feedback accumulates.

\subsection{Behavioral Route Diagnostics}

\begin{table}[t]
\centering
\caption{Observable routing behavior on the 2025 evaluation vintage. Fixed \rel{} avoids the fitted router's overuse of graph/search behavior; the self-adjusting rule is more conservative on this vintage and shifts most targets to historical analog behavior.}
\label{tab:behavior}
\begin{tabular}{lrrrr}
\toprule
Router & Historical & AIA-repro & Search & Graph \\
\midrule
\rel{} fixed & 589 & 488 & 0 & 0 \\
\rel{} self-adjusting & 1004 & 73 & 0 & 0 \\
\tax{} & 588 & 489 & 0 & 0 \\
\fitroute{} & 196 & 11 & 288 & 582 \\
\bottomrule
\end{tabular}
\end{table}

Table~\ref{tab:behavior} reports the behavioral intervention directly. The fixed \rel{} rule routes 589 of 1077 targets to historical analog behavior and 488 to conservative reproduced-baseline behavior, nearly matching the hand taxonomy without source-name rules. The self-adjusting rule, after seeing earlier 2025 vintages, routes 1004 targets to historical analog behavior and only 73 to the conservative baseline on this vintage; this explains why it is better calibrated but not better in Brier on Table~\ref{tab:main}. By contrast, the fitted router routes 870 targets to graph/search behavior and performs worse. This supports the behavioral interpretation of the main result: the failure mode is not simply poor probability calibration, but unstable behavioral mechanism choice under transfer.

\subsection{Multi-Vintage Stress Test}

\begin{table}[t]
\centering
\caption{Mean Brier score across 16 later LLM-written ForecastBench vintages. The self-adjusting reliability router refits thresholds in walk-forward order using only previously resolved vintages. Lower is better.}
\label{tab:multivintage}
\begin{tabular}{lccc}
\toprule
System & Mean Brier $\downarrow$ & Vintage wins & Median rank \\
\midrule
\rel{} self-adjusting & \textbf{0.1839} & 7/16 & 2.0 \\
\hist{} & 0.1841 & 2/16 & 3.0 \\
\search{} & 0.1848 & 1/16 & 4.0 \\
\tax{} & 0.1863 & 4/16 & 3.5 \\
\rel{} fixed & 0.1867 & 1/16 & 4.0 \\
\fitroute{} & 0.1940 & 1/16 & 6.0 \\
\aialocal{} & 0.2054 & 0/16 & 7.0 \\
\bottomrule
\end{tabular}
\end{table}

Table~\ref{tab:multivintage} extends the evaluation to 16 LLM-written vintages from 2025-03-02 through 2025-12-07. This broader test changes the interpretation of \rel{}: fixed thresholds are useful but brittle, while walk-forward self-adjustment yields the best mean Brier among our deterministic systems. The self-adjusting router improves mean Brier from 0.1867 for the frozen rule to 0.1839, wins 7 of 16 vintages, and beats the frozen rule on 10 of 16 vintages. A vintage-level paired bootstrap gives a mean Brier difference of $-0.0029$ against the frozen rule with 95\% interval $[-0.0056,-0.0006]$. The gain remains modest: \hist{} is close at 0.1841, and the bootstrap interval against \hist{} overlaps zero. This supports a behavioral rather than leaderboard-style interpretation: routing helps when it adapts under constraints, but many targets are already well served by simple structured analogs or deterministic search evidence.

Threshold and route traces explain what self-adjustment changes behaviorally. After early vintages, the rule becomes stricter about source-level historical coverage, raises market-prior sharpness thresholds, and mostly disables search fallback when prior vintages do not support it; the full per-vintage rule trace is included in the reproducibility artifacts. Across vintages, single-threshold perturbations change 4.3\% of routes on average and change Brier by 0.0007 on average, although the most disruptive perturbations can change up to 21\% of routes on some vintages. The intervention is therefore not merely changing probabilities; it is changing which behavioral mechanism is allowed to control the forecast, while leaving a visible audit trail of rule drift.

\subsection{Source-Slice Diagnostics}

\begin{table}[t]
\centering
\caption{Source-level Brier score on the 2025 evaluation vintage. Parentheses show the number of targets. The best mechanism depends strongly on source family, and fixed versus self-adjusting routing make different source-level tradeoffs.}
\label{tab:source}
\resizebox{\linewidth}{!}{%
\begin{tabular}{lccccc}
\toprule
Source & \rel{} fixed & \rel{} self-adj. & \tax{} & \hist{} & \aialocal{} \\
\midrule
ACLED (200) & \textbf{0.1112} & \textbf{0.1112} & \textbf{0.1112} & \textbf{0.1112} & 0.1543 \\
DBnomics (190) & \textbf{0.2330} & 0.2470 & \textbf{0.2330} & 0.2470 & \textbf{0.2330} \\
FRED (196) & \textbf{0.2292} & \textbf{0.2292} & \textbf{0.2292} & \textbf{0.2292} & 0.2564 \\
Infer (5) & \textbf{0.0391} & 0.0548 & \textbf{0.0391} & 0.0548 & \textbf{0.0391} \\
Manifold (21) & \textbf{0.0347} & 0.0370 & \textbf{0.0347} & 0.0377 & \textbf{0.0347} \\
Metaculus (11) & 0.2321 & 0.2111 & 0.2321 & \textbf{0.2072} & 0.2321 \\
Polymarket (70) & 0.0170 & 0.0205 & \textbf{0.0167} & 0.0209 & \textbf{0.0167} \\
Wikipedia (192) & \textbf{0.1821} & \textbf{0.1821} & \textbf{0.1821} & \textbf{0.1821} & 0.2006 \\
YFinance (192) & \textbf{0.2489} & 0.2508 & \textbf{0.2489} & 0.2508 & \textbf{0.2489} \\
\bottomrule
\end{tabular}}
\end{table}

Table~\ref{tab:source} explains why routing helps and why the two \rel{} variants differ. Historical analogs dominate for ACLED, FRED, and Wikipedia. The reproduced multi-agent baseline is stronger for DBnomics, YFinance, and several market-family sources. Fixed \rel{} recovers almost the same slice behavior as \tax{} from reliability features rather than source names. Self-adjusting \rel{} improves some small market-like slices by using historical behavior more often, but loses on DBnomics and YFinance on this vintage by over-shifting toward historical analogs. This is the single-vintage cost of the adaptation that improves average multi-vintage performance.

\subsection{Ablations and Negative Results}

\fitroute{} directly fits the best mechanism per source on the 2024 vintage, so it is intentionally more label-dependent than \tax{} or \rel{}. This diagnostic performs worse than using the source taxonomy. It selects evidence graph for DBnomics and YFinance because those choices win on the calibration vintage, but the same choices fail on the 2025 vintage. This yields Brier 0.1949, worse than \hist{} and \search{}. The result shows that behavioral routing is useful only when the routing rule is constrained by reliability structure or validated across vintages.

\begin{table}[t]
\centering
\caption{Routing and calibration ablations on the 2025 evaluation vintage. Oracle rows use 2025 labels and are diagnostic upper bounds rather than deployable systems. Lower is better for all metrics.}
\label{tab:ablations}
\begin{tabular}{lccc}
\toprule
System & Brier $\downarrow$ & Log score $\downarrow$ & Cal. error $\downarrow$ \\
\midrule
Oracle question router & \textit{0.1196} & \textit{0.3822} & 0.2369 \\
Oracle source router & \textit{0.1844} & \textit{0.5377} & 0.0634 \\
\tax{} & \textbf{0.1846} & \textbf{0.5393} & 0.0650 \\
\rel{} fixed & \textbf{0.1846} & \textbf{0.5394} & 0.0649 \\
\hist{} & 0.1876 & 0.5460 & \textbf{0.0240} \\
\rel{} self-adjusting & 0.1876 & 0.5461 & \textbf{0.0232} \\
\tax{} + global calibration & 0.1882 & 0.5531 & 0.0670 \\
\fitroute{} & 0.1949 & 0.5651 & 0.0797 \\
\tax{} + source calibration & 0.2004 & 0.5798 & 0.1025 \\
\aialocal{} & 0.2009 & 0.5808 & 0.1200 \\
\bottomrule
\end{tabular}
\end{table}

Table~\ref{tab:ablations} shows two additional negative results. First, post-hoc Platt-style calibration fitted on the 2024 vintage does not repair the routed system: global calibration worsens Brier from 0.1846 to 0.1882, and source-conditional calibration worsens it to 0.2004. This suggests that the calibration mismatch is not a single global slope/intercept problem and that source-level calibration overfits when each slice is small. Second, the oracle source router reaches 0.1844 on the 2025-10-26 vintage, only 0.0002 Brier better than the fixed \rel{} rule on that diagnostic split. Thus, much of the single-vintage source-level headroom is already captured by reliability routing, while the multi-vintage results show that threshold adaptation matters for transfer. The much stronger oracle question router shows that fine-grained per-question routing could matter, but using evaluation labels makes this only an upper bound.

\section{Related Work}

Forecasting has a long tradition of evaluating probabilistic predictions with proper scoring rules \citep{brier1950verification,gneiting2007strictly}. Human forecasting research shows that expert forecasters can be systematically trained and evaluated \citep{tetlock2015superforecasting}, while prediction markets aggregate dispersed information through market prices \citep{hanson2003combinatorial}. ForecastBench provides a dynamic benchmark for comparing AI, public, and expert forecasts on future-resolving questions \citep{karger2024forecastbench}. AIA Forecaster is a closely related LLM forecasting system that combines agentic search, supervisor aggregation, and calibration \citep{alur2025aia}; we use an AIA-style reproduced baseline as one comparator in our harness. Our work focuses on a different axis: isolating mechanism choice and evaluating it as an observable agent behavior. This connects forecasting accuracy to a behavioral question: when should an agent retrieve, reason, defer to an external prior, or rely on structured historical evidence?

\section{Limitations}

Our study is intentionally component-level. Our AIA-style reproduced baseline does not use the same search stack, prompts, models, or calibration procedure as the reported AIA Forecaster system, so it should be read as a baseline for comparison rather than a reproduction claim. The main single-vintage gain is modest, and the paired bootstrap interval against the strongest single baseline slightly overlaps zero. The multi-vintage self-adjusting gain is also narrow: \rel{} has the best mean Brier in our deterministic harness, but \hist{} is close and several vintages are still won by simple baselines or the hand taxonomy. \rel{} replaces source-name routing with feature thresholds, but those thresholds are selected from a human-designed rule family; future work should validate the same adaptation procedure with nested development splits and external forecasting datasets. Our behavioral analysis also observes only mechanism selection and route-level features, not richer interaction behavior such as search-query refinement, tool-use trajectories, or multi-agent disagreement dynamics. Finally, \rel{} improves Brier and log score on some vintages but often worsens calibration error, and our train-fitted global and source-conditional calibration attempts both fail to transfer. Downstream users who require calibrated probabilities should therefore treat conditional calibration as an unsolved component rather than a solved add-on.

\section{Conclusion}

Forecasting agents should not reason uniformly across all questions. In our ForecastBench transfer study, historical analogs, market/crowd-style baselines, and search/graph reasoning have different source-dependent error profiles. A feature-based reliability router can improve on individual vintages, and walk-forward self-adjustment improves mean multi-vintage Brier while remaining more constrained than directly fitting per-source winners. The practical lesson is therefore not that routing alone solves forecasting. It is that behavior choice is itself an object of evaluation: before adding more reasoning, a forecasting agent should estimate which evidence source deserves control, update that estimate only from resolved evidence, and spend future complexity on within-source forecasting, richer behavioral traces, and robust conditional calibration.

\section*{Reproducibility Statement}

We provide an reproducibility artifact at \url{https://github.com/louiswang524/forcastagent}. The artifact includes the deterministic forecasting-agent harness, cached public ForecastBench inputs, smoke tests, and scripts for rerunning the fixed-transfer, calibration, source-slice, threshold-stability, and walk-forward self-adjustment experiments reported in this paper. No API key is required for the deterministic results.

\section*{Ethics Statement}

This work studies probabilistic forecasting systems. Such systems can support better decision-making, but overconfident or poorly calibrated forecasts may cause harm if used without human oversight. We therefore report calibration error, negative transfer results, and limitations alongside accuracy metrics.

\bibliography{references}
\bibliographystyle{colm2026_conference}

\end{document}